\documentclass{article}

\IfFileExists{neurips_2025.sty}{
  \usepackage{neurips_2025}
}{
  \usepackage[a4paper,margin=1in]{geometry}
}

\usepackage[T1]{fontenc}
\usepackage[utf8]{inputenc}
\usepackage{lmodern}
\usepackage{microtype}
\usepackage{booktabs}
\usepackage{longtable}
\usepackage{array}
\usepackage{amsmath,amssymb}
\usepackage{graphicx}
\usepackage{subcaption}
\usepackage{enumitem}
\usepackage{float}
\usepackage[section]{placeins}
\usepackage{hyperref}
\usepackage{url}
\usepackage{natbib}
\usepackage{tikz}
\usetikzlibrary{arrows.meta,positioning,shapes.geometric}

\graphicspath{{figures/}}

\hypersetup{colorlinks=true,linkcolor=blue,citecolor=blue,urlcolor=blue}
\title{A Control Architecture for Training-Free Memory Use}
\author{
\parbox[t]{0.42\textwidth}{\centering
Yanzhen Lu\\[0.4ex]
\small Shanghai Qizhi Institute\\[0.2ex]
\small \texttt{luyanzhen@sqz.ac.cn}
}\hfill
\parbox[t]{0.42\textwidth}{\centering
Muchen Jiang\\
\small Nanjing Whale Cloud\\[0.4ex]
\small \texttt{jiang.muchen2@iwhalecloud.com}\\[0.2ex]
}\\[6.0ex]
\parbox[t]{0.42\textwidth}{\centering
Zhicheng Qian\\[0.4ex]
\small Southeast University\\[0.2ex]
\small \texttt{220232282@seu.edu.cn}
}\hfill
\parbox[t]{0.42\textwidth}{\centering
Xingyu Zhou\\[0.4ex]
\small Xiamen University\\[0.2ex]
\small \texttt{33520241153320@stu.xmu.edu.cn}
}
}
\date{}

\newcommand{\method}{TAG}

\newcommand{\hm}{\text{Help}-\text{Hurt}}
\newcommand{\acc}{\text{Acc}}

\begin{document}
\maketitle
\begin{abstract}
Prompt-injected memory can improve reasoning without updating model weights, but it also creates a control problem: retrieved content helps only when it is applied in the right state.
We study this problem in a strict training-free setting and formulate it as \emph{applicability control}: when to trigger a memory-assisted second pass, when to trust it, and how to maintain the memory bank over time.
Our method combines uncertainty-based routing, confidence-based selective acceptance, bank selection across rule and exemplar memory, and evidence-based governance of the memory bank over time.
Under a locked training-free protocol with compute-matched controls, it improves two core arithmetic benchmarks by +7.0 points on SVAMP and +7.67 points on ASDiv over baseline.
The same architecture also transfers to QA and agent benchmarks with smaller positive effects and shows the same positive direction on a second checkpoint for the main arithmetic tasks.
On arithmetic, the main empirical pattern is that the control architecture, rather than raw memory exposure, drives the improvements on SVAMP and ASDiv. Mechanistically, confidence separates helpful from harmful rule-bank interventions, and under fixed retrieval the repair-versus-corrupt difference localizes to rows whose retrieved set actually contains the edited entries.
\end{abstract}
\section{Introduction}
\label{sec:intro}
\begin{figure}[H]
\centering
\includegraphics[width=\linewidth]{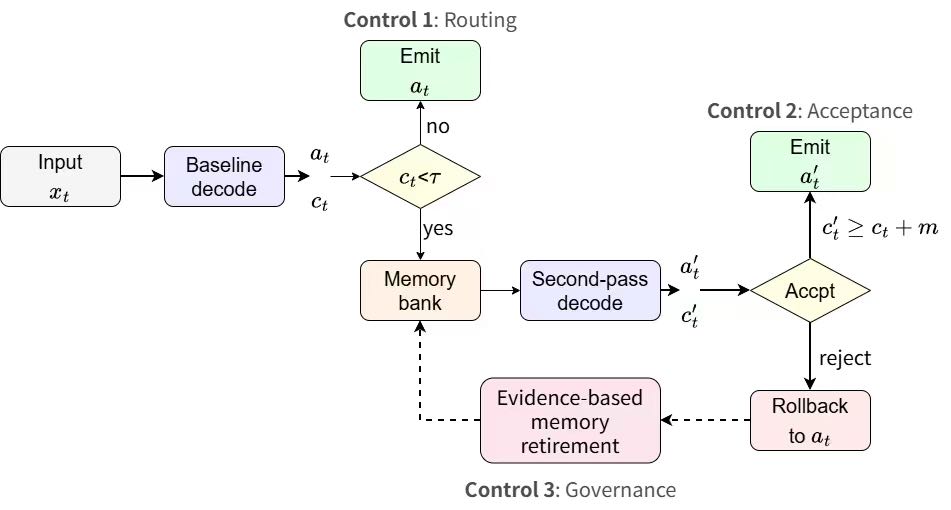}
\caption{\textbf{\method{} decision flow.} At each step, the baseline action is decoded and its confidence assessed. Low-confidence steps are routed to memory retrieval and a second-pass decode. The second-pass output is accepted only if it passes both a confidence-margin check and structural guards; otherwise the system rolls back to the baseline action. Over time, persistently harmful memory entries are retired.}
\label{fig:method}
\end{figure}

Training-free context optimization treats prompts and memory as the optimization object instead of model parameters. Its attraction is practical: no weight updates, no base-model drift, and direct deployment on top of existing systems. Prompt-injected memory is a natural instance of this idea, but it works only when retrieved content is applied under the right conditions. A rule or exemplar that is useful in one state can be irrelevant or harmful in another. This makes prompt memory a control problem as much as a retrieval problem. In addition to choosing what to retrieve, the system must decide when memory should be queried, which bank should be exposed, and whether the memory-conditioned answer should override the baseline. We therefore study prompt memory as an applicability-control problem under locked evaluation.
\paragraph{The applicability-control challenge.}
Most prior work focuses on ``what to retrieve'' but gives less attention to ``whether to apply retrieved content at all.'' Inappropriate application both hurts quality and wastes compute. In our locked protocol, a compute-matched retry produces no lift on the strict reasoning benchmarks, and stronger retrieval-only or non-memory controls remain below the best gated memory policy on SVAMP and ASDiv. The main issue is therefore applicability control under fixed compute, not raw retrieval volume.
We propose \method{} as a training-free control stack with four coupled pieces:
(1)~\emph{uncertainty-based routing};
(2)~\emph{confidence-based selective acceptance};
(3)~\emph{bank selection} across rule and exemplar memory;
(4)~\emph{evidence-based governance} of the memory bank over time.
Figure~\ref{fig:method} illustrates the decision flow.
\paragraph{Contributions.}
\begin{enumerate}[leftmargin=1.5em,itemsep=2pt]
\item We formulate prompt-memory applicability as a control problem and evaluate it under locked, compute-matched comparison.
\item We introduce a training-free control architecture that combines uncertainty-gated routing, confidence-based selective acceptance, bank selection across rule and exemplar memory, and explicit memory governance.
\item We show that the main arithmetic gains are not retrieval-only and not retry-only: on SVAMP and ASDiv, the control layer outperforms weaker-control and non-memory alternatives built on the same prompt-memory substrate.
\item We provide mechanism evidence from bank asymmetry, bank-specific confidence separability, and fixed-retrieval counterfactual replay, which isolates retrieval drift and localizes the residual content-edit signal to edited-entry exposures.
\item We report supporting transfer and robustness evidence on QA, agent, second-checkpoint, and same-solver memory-quality comparisons.
\end{enumerate}
\paragraph{Scope.}
This paper focuses on the control layer rather than on designing a new retriever. Our goal is to separate memory content from memory-use policy under a locked protocol and identify robust operating points for memory use.
\section{Related Work}
\label{sec:related}
\paragraph{Selective prediction and abstention.}
Selective classification~\citep{geifman2017selective,kamath2020selective}, LLM self-assessment~\citep{kadavath2022language,lin2022teaching,xiong2024can}, semantic uncertainty~\citep{kuhn2023semantic}, and calibration~\citep{guo2017calibration,platt1999probabilistic} all motivate our use of confidence as a control signal. \method{} turns that idea into a routing gate inside a memory-augmented pipeline.
\paragraph{Retrieval-augmented generation and control.}
RAG~\citep{lewis2020retrieval}, REALM~\citep{guu2020realm}, Self-RAG~\citep{asai2024selfrag}, CRAG~\citep{yan2024corrective}, FLARE~\citep{jiang2023active}, and uncertainty-triggered retrieval~\citep{yao2025seakr} study when retrieval should occur or how retrieval quality should improve. Our focus is more intervention-centric: whether and when retrieved memory should override the baseline action, with explicit rollback and retirement.
\paragraph{Memory-augmented agents.}
Generative agents~\citep{park2023generative}, Voyager~\citep{wang2023voyager}, Reflexion~\citep{shinn2023reflexion}, CLIN~\citep{majumder2023clin}, ExpeL~\citep{zhao2024expel}, and MemGPT~\citep{packer2023memgpt} show how memory can support long-horizon behavior. \method{} complements this line by governing \emph{when} stored knowledge should be applied.
\paragraph{Prompt optimization.}
Prompt-optimization methods such as DSPy~\citep{khattab2023dspy}, APE~\citep{zhou2023large}, OPRO~\citep{yang2024large}, and self-consistency~\citep{wang2023selfconsistency} improve content or reasoning quality; \method{} instead governs when memory content is applied.
\paragraph{Training-free experience libraries.}
Recent training-free prompt-memory systems, including Training-Free GRPO-style experience libraries (arXiv:2510.08191), collect natural-language experiences and reinject them at test time. Our paper is a control-layer study on top of that same substrate: when should memory be exposed, when should it override the baseline, and when should harmful memories be retired?
\section{Method}
\label{sec:method}
\paragraph{Problem formulation.}
At step~$t$, a baseline policy proposes action~$a_t$ with confidence~$c_t$; if retrieval is triggered, a memory-conditioned second pass proposes~$a_t'$ with confidence~$c_t'$. We optimize a quality--cost trade-off under training-free constraints:
\begin{equation}
\max_{\pi\in\Pi_{\text{tf}}} \;\mathbb{E}[Q(\pi)] - \lambda\,\mathbb{E}[C(\pi)],
\label{eq:objective}
\end{equation}
where $Q$ is task quality, $C$ is calls per query, and $\Pi_{\text{tf}}$ forbids parameter updates.
\paragraph{Applicability as latent intervention value.}
We separate \emph{retrieval exposure} from \emph{useful applicability}. Let $e_t$ denote the retrieved memory item or bank shown at step~$t$, and let $A_t(e_t)\in\{0,1\}$ indicate whether that memory is actually valid for the current local decision. Define the incremental utility of committing the memory-conditioned action as
\begin{equation}
\Delta_t(e_t)=u_t(a_t';e_t)-u_t(a_t),
\label{eq:delta}
\end{equation}
where $u_t$ is task utility at the current step or example. In this formulation, the sign of $\Delta_t(e_t)$ is governed mainly by whether $A_t(e_t)=1$.
\paragraph{Shared control target.}
All three controls approximate the sign of a single latent value. Let
\begin{equation}
V_t(e_t)=\mathbb{E}[\Delta_t(e_t)-\lambda k_t\mid x_t,e_t],
\label{eq:shared_value}
\end{equation}
where $k_t$ is the extra cost of intervention. Routing estimates whether exposure is worth buying; acceptance estimates whether the realized candidate should override the baseline; retirement estimates whether an entry has negative average value across episodes. This provides an intuition for why route/accept/retire can be analyzed within one intervention-value view.
\paragraph{Stagewise view.}
The same intervention-value view separates three kinds of control error: querying in the wrong states, committing the wrong second-pass proposals, and retaining persistently harmful entries. We use this decomposition only as a guide for organizing the controller and the empirical comparisons; the formal interpretation appears in Appendix~\ref{app:theory}.
\paragraph{Why exposure alone is insufficient.}
An exposure-only policy conditions on retrieved identity but not on whether the current state actually satisfies that memory's validity conditions. If $\Pr(A_t=1\mid x_t,e_t)$ varies across states that retrieve the same item, exposure averages together positive and negative regimes and can drive $V_t(e_t)$ toward zero or below it. A useful controller therefore has to condition on state-local evidence, not on retrieval match alone.
\paragraph{Decision factorization.}
A generic controller would choose
\begin{equation}
d_t^\star=\arg\max_{d\in\mathcal{D}} \mathbb{E}\!\left[\Delta_t(d)-\lambda k_t(d)\mid x_t\right],
\label{eq:bayes}
\end{equation}
where $d$ ranges over baseline-only, retrieve-and-commit, or retrieve-and-reject actions and $k_t(d)$ is the incremental compute cost. \method{} approximates this decision by factorizing it into three simpler controls: whether to query memory, whether to accept the resulting candidate, and whether the corresponding memory entry should remain active in future episodes.
\subsection{Unified Control Loop}
\method{} is a single intervention policy rather than three unrelated heuristics. Each step asks three questions in sequence: should memory be consulted, should the memory-conditioned proposal be trusted, and should the responsible entry remain active in future episodes.

\begin{table}[H]
\centering
\small
\begin{tabular}{p{0.96\linewidth}}
\toprule
\textbf{Algorithm 1: \method{} inference and fit-stage governance}\\
\midrule
\textbf{Input:} query or agent state $x_t$, frozen memory bank $\mathcal{M}$, routing threshold $\tau$, acceptance margin $m$, deterministic guards $g_t$, and optional episode budget.\\
1.~Decode a baseline action $a_t$ and confidence $c_t$.\\
2.~If $c_t \ge \tau$, emit $a_t$ and do not retrieve memory at this step.\\
3.~Otherwise retrieve candidate memory from $\mathcal{M}$, run a memory-conditioned second pass, and obtain $(a_t', c_t')$.\\
4.~Accept $a_t'$ only if it exceeds the baseline by margin $m$ and passes the structural guard conjunction $g_t$; otherwise roll back to $a_t$.\\
5.~On fit/dev data only, attach paired utility evidence from the accepted or rejected intervention to the retrieved memory entry or entries.\\
6.~After fit/dev evaluation, retire entries whose upper confidence bound on mean utility is below zero; freeze the resulting bank and all selected thresholds before test.\\
\bottomrule
\end{tabular}
\caption{\textbf{Execution summary of \method{}.} The training-free constraint applies throughout: the model weights never change, and only prompts, routing decisions, and fit-time memory-bank membership are updated.}
\label{tab:tag_algorithm}
\end{table}
\subsection{Control 1: Uncertainty-Based Routing}
Route to memory retrieval only when baseline confidence is below a threshold:
\begin{equation}
\text{route}_t = \mathbf{1}[c_t < \tau].
\label{eq:route}
\end{equation}
This suppresses unnecessary second passes when the baseline is already confident, directly reducing~$C$. Operationally, routing approximates the event
\begin{equation}
\mathbb{E}[\Delta_t(e_t)-\lambda k_t \mid x_t, c_t] > 0,
\label{eq:route_gain}
\end{equation}
using $c_t$ as a proxy for residual headroom in the baseline policy. For reasoning tasks we use answer confidence; for agents we use step-level action confidence. The threshold $\tau$ is selected on fit/dev data and frozen for test.
\subsection{Control 2: Guarded Acceptance with Rollback}
After second-pass decoding, acceptance requires both a confidence improvement and structural compatibility:
\begin{equation}
\text{accept}_t = \mathbf{1}[c_t' \ge c_t + m]\cdot \mathbf{1}[g_t=1],
\label{eq:accept}
\end{equation}
where $m$ is a confidence margin and $g_t$ is a deterministic guard conjunction.
\begin{equation}
g_t = g_t^{\text{format}} \land g_t^{\text{valid}} \land g_t^{\text{progress}} \land g_t^{\text{contract}},
\label{eq:guard}
\end{equation}
where inactive terms are treated as $1$ when they are not relevant to the current benchmark.
If rejected, the system rolls back to~$a_t$. Routing decides whether memory is \emph{queried}; acceptance decides whether the observed intervention has enough evidence to overcome baseline inertia. For reasoning, guards mostly verify answer format and schema consistency; for agents, they emphasize action validity, observable progress, and contract adherence.
\subsection{Control 3: Evidence-Based Retirement}
Memory entries accumulate paired utility evidence across evaluations.
An entry is retired when its upper confidence bound on mean utility falls below zero, following a UCB-style rule with Hoeffding radius~\citep{auer2002finite,hoeffding1963probability}.
This changes memory policy state without model updates. Each routed intervention contributes paired utility evidence relative to the baseline proposal. Persistently harmful entries are removed before frozen test evaluation.
\subsection{Fit-Stage Governance and Frozen Test-Time Policy}
\method{} uses a two-stage boundary. During fit/dev, we select thresholds, margins, budgets, policy family, and governance iteration, and update memory-bank membership through retirement. During test, all choices are frozen.
\subsection{Budget Discipline}
For agents, per-episode route-budget caps and cooldown periods limit~$C$ directly. Budgeting complements routing: the threshold controls \emph{which} steps are eligible, while the budget controls \emph{how many} are served per episode.
\paragraph{Implications for evaluation.}
This framing yields three empirical comparisons: $\Delta_{\mathrm{compute}} = Q(\pi_{\mathrm{retry}})-Q(\pi_0)$, $\Delta_{\mathrm{exposure}} = Q(\pi_{\mathrm{gate}})-Q(\pi_{\mathrm{exp}})$, and $\Delta_{\mathrm{fixed\text{-}cf}} = \mathbb{E}[Y(x,R(x,M),\widetilde m)-Y(x,R(x,M),m)]$. We use these quantities to organize the experiments rather than to claim a new theoretical algorithm.
\section{Experimental Setup}
\label{sec:setup}
\paragraph{Benchmarks.}
\emph{Reasoning:} SVAMP~\citep{patel2021svamp}, ASDiv~\citep{miao2020asdiv}, and MultiArith~\citep{roy2015multiarith} under a strict fit$\rightarrow$test protocol with Qwen3-0.6B~\citep{yang2025qwen3}.
\emph{Agents:} WebShop~\citep{yao2022webshop} ($n\!=\!900$, pooled over 3 seeds) and ScienceWorld~\citep{wang2022scienceworld} ($n\!=\!600$, pooled over 3 seeds), both with Qwen3-8B~\citep{yang2025qwen3}.
\paragraph{Memory Bank Construction.}
For reasoning tasks, memory banks contain rules (general mathematical principles, $n=50$ per dataset) and exemplars (annotated problem-solution pairs, $n=100$ per dataset). For agents, memory entries consist of error-recovery hints and solution fragments ($n=200$ for WebShop, $n=150$ for ScienceWorld). Banks are built from train-side or fit-side resources only, then frozen before test evaluation; representative protocol notes appear in Appendix~\ref{app:protocol}. For retrieval, we use all-MiniLM-L6-v2 embeddings with similarity thresholds ($>0.6$ for reasoning, $>0.7$ for agents).
\paragraph{Confidence instantiation.}
For reasoning tasks, confidence is the mean token log-probability of the decoded answer (\texttt{answer\_logprob\_mean}).
For agents, step-level confidence is the mean token log-probability of the generated action text.
We use confidence as a gating signal, not as a calibrated probability; see Appendix~\ref{app:confidence} for sweep and calibration diagnostics.
\paragraph{Evaluation protocol.}
All comparisons are \emph{paired} at the same example or goal index. Thresholds, margins, budgets, policy-family choice, and retained governance iteration are selected on fit/dev splits and \emph{frozen} before test reporting. Governance updates are permitted only during fit. We report multi-seed pooled results with effect sizes, 95\% bootstrap confidence intervals, and McNemar paired tests. Primary cost comparisons use normalized calls/query; token and wall-time metadata are reported only where coverage is sufficient.
\paragraph{Metrics.}
We report accuracy gain ($\Delta$acc), score gain ($\Delta$score), call reduction ($\Delta$calls), and Help--Hurt counts, plus intervention diagnostics such as routed steps/query, accepted steps/query, invalid steps/query, and trace-level reject reasons for agents.
\paragraph{Baselines.}
(1)~\textbf{No-memory baseline}: single-pass decoding.
(2)~\textbf{Retry}: compute-matched second pass without memory.
(3)~\textbf{Fixed-budget retrieval}: a training-free memory baseline that retrieves up to $k\!=\!2$ per episode with no guards or retirement.
(4)~\textbf{Always-retrieve}: a training-free memory baseline that retrieves at every step with no rollback (ReAct-like).
These are our direct training-free comparators under the same checkpoint and prompt-memory substrate. Appendix~\ref{app:baselines} reports supplementary same-checkpoint controls spanning compute-only and inference-only alternatives (retry, reflection, self-consistency, and verifier filtering) and retrieval-exposure variants that change memory use without rollback. Appendix~\ref{app:baselines} also reports a separate same-solver memory-quality comparison against official TFGRPO memories on AIME24.
\section{Results}
\label{sec:results}
\subsection{Arithmetic Main Results}
\label{sec:results_reasoning}
The strict main table shows that the benefit of prompt memory depends on the control policy. \method{} produces substantial gains on SVAMP and ASDiv, a smaller non-significant lift on MultiArith, and a flat compute-matched retry control on all three datasets.
\begin{table}[H]
\centering
\small
\begin{tabular}{lcccccc}
\toprule
Dataset & Baseline & Retry & \method{} & Oracle & $\Delta$acc & $p$ \\
\midrule
SVAMP   & 74.0 & 74.0 & 81.0 & 84.5 & +7.0 \tiny{[+4.3, +9.8]} & $9.7{\times}10^{-7}$ \\
ASDiv   & 77.5 & 77.5 & 85.2 & 87.0 & +7.7 \tiny{[+5.5, +10.0]} & $3.8{\times}10^{-11}$ \\
MultiArith & 87.8 & 88.0 & 89.1 & 92.8 & +1.3 \tiny{[$-$0.2, +2.8]} & 0.143 \\
\bottomrule
\end{tabular}
\caption{\textbf{Reasoning results} (multi-seed means, \%). \method{} shows the fit-selected policy per dataset. Oracle is the paired upper bound from accepting only empirically helpful interventions. $\Delta$acc is vs.\ baseline with 95\% CI in brackets; $p$ is McNemar's test. Retry is compute-matched. Full family and significance ledger are in the Appendix.}
\label{tab:reasoning_main}
\end{table}

Table~\ref{tab:reasoning_main} reports the fit-selected frozen policy for each dataset. The best policy changes by dataset: SVAMP prefers a stronger rule-centric configuration, ASDiv prefers exemplar-first routing, and MultiArith uses a more conservative rule policy. Retry remains flat. The paired gains on the favorable datasets are large and stable: SVAMP dual vs.\ baseline yields $\Delta\acc=+0.0700$ with $p=9.67\times10^{-7}$, and ASDiv cascade exemplar$\rightarrow$rule yields $\Delta\acc=+0.0767$ with $p=3.80\times10^{-11}$.

\subsection{Stronger Alternatives}
\label{sec:results_retrieval_only}
On SVAMP and ASDiv, gated memory remains above same-checkpoint retrieval-control and inference-only alternatives.
\begin{table}[H]
\centering
\small
\begin{tabular}{lccccc}
\toprule
Dataset & Best alternative & Baseline & Alternative & Gated memory & $\Delta$(Gated$-$Alt.) \\
\midrule
SVAMP & verifier & 74.0 & 75.7 & 77.8 & +0.0216 \\
ASDiv & always-retrieve & 77.5 & 79.2 & 81.2 & +0.0200 \\
MultiArith & always-retrieve & 87.8 & 89.4 & 89.6 & +0.0019 \\
\bottomrule
\end{tabular}
\caption{\textbf{Retrieval-control and inference-only alternatives.} Always-retrieve uses the same prompt-memory substrate without applicability control. Positive $\Delta$(Gated$-$Alt.) means the control stack stays above the best listed retrieval-control or inference-only alternative on that dataset.}
\label{tab:retrieval_only_main}
\end{table}
Within the same memory substrate, the gated policy stays above always-retrieve and verifier-based alternatives by about two points on SVAMP and ASDiv. Appendix~\ref{app:baselines} reports the broader same-checkpoint comparison, including retry, reflection, self-consistency, verifier filtering, and reranked retrieval-only. The appendix also reports a separate same-solver memory-quality comparison against official TFGRPO memories on AIME24.

\subsection{Component Isolation}
\label{sec:results_components}
The stack is not a single routing trick; different datasets need different parts of the control loop.
\begin{table}[H]
\centering
\small
\begin{tabular}{lcccccc}
\toprule
Dataset & Gate-only & Choose & Multibank & $\Delta$(Gate$-$Base) & $\Delta$(Choose$-$Gate) & $\Delta$(Multi$-$Choose) \\
\midrule
SVAMP & 80.0 & 80.0 & 80.3 & +0.0733 & +0.0000 & +0.0033 \\
ASDiv & 84.7 & 83.7 & 85.7 & +0.0633 & -0.0100 & +0.0200 \\
MultiArith & 83.3 & 92.6 & 91.5 & -0.0889 & +0.0926 & -0.0111 \\
\bottomrule
\end{tabular}
\caption{\textbf{Component isolation.} These are control-stack ablations on the same prompt-memory substrate. Favorable datasets need more than routing alone, but not in the same way.}
\label{tab:component_main}
\end{table}
Because every column uses the same underlying memory substrate, the differences isolate which control stage is doing work. On SVAMP, gate-only already helps and multibank composition adds a further gain. On ASDiv, gate-only is strong, selective acceptance within the best single bank is weaker, and multibank composition then recovers the strongest result. Across datasets, the effective control stage depends on the bank and benchmark.
Appendix~\ref{app:baselines} reports the full locked component table and two associated patterns. First, compute-matched retry is flat on SVAMP and ASDiv within the same evaluation setting, so the gains are not a generic second-pass artifact. Second, bank selection matters because different banks benefit from selective acceptance in different ways.

\subsection{Cross-Domain QA Support}
\label{sec:results_crossdomain}
\begin{table}[H]
\centering
\scriptsize
\begin{tabular}{lccccc}
\toprule
Dataset & Base & Retr. & Retry & Gate & $\Delta$(G$-$R) \\
\midrule
OpenBookQA & 50.5 & 48.2 & 50.5 & 51.8 & +0.0367 \\
ARC-Easy & 66.3 & 70.2 & 66.3 & 72.5 & +0.0233 \\
ARC-Challenge & 25.5 & 27.5 & 25.2 & 27.8 & +0.0033 \\
\bottomrule
\end{tabular}
\caption{\textbf{Cross-domain QA.} OpenBookQA shows recovery relative to retrieval exposure, ARC-Easy shows a larger gated-over-retrieval gain, and ARC-Challenge shows a smaller difference.}
\label{tab:cross_domain_main}
\end{table}
OpenBookQA recovers and slightly improves over baseline after retrieval alone falls below baseline. ARC-Easy shows a larger gated-over-retrieval gain, while ARC-Challenge shows a smaller difference.

\subsection{Agent-Benchmark Support}
\label{sec:results_agents}
\begin{table}[H]
\centering
\scriptsize
\begin{tabular}{lccccc}
\toprule
Benchmark & $n$ & $\Delta$acc & Help$-$Hurt & $\Delta$score & $\Delta$calls \\
\midrule
WebShop & 900 & +0.0144 {\tiny[+0.0067,+0.0233]} & +13 & +0.2525 & $-$0.5500 \\
ScienceWorld & 600 & +0.0017 {\tiny[+0.0000,+0.0050]} & +1 & +1.0633 & $-$0.9117 \\
\bottomrule
\end{tabular}
\caption{\textbf{Agent-benchmark support.} Pooled paired comparisons for the selected agent policies. WebShop shows a clean accuracy/help-hurt gain; ScienceWorld is closer to a score-and-cost improvement than a raw-accuracy win.}
\label{tab:agent_support_main}
\end{table}
On WebShop, the selected autoretired policy improves accuracy by $+1.44$ points with 14 helps against 1 hurt, raises mean score by $+0.2525$, and reduces calls/query by $0.55$ relative to the corresponding no-memory reference policy. On ScienceWorld, the selected adaptive policy is close to accuracy-neutral but still improves score and cost, with a paired gain of $+0.17$ points, Help$-$Hurt $=+1$, score gain $+1.0633$, and $0.91$ fewer calls/query than the corresponding frozen reference policy. Appendix~\ref{app:significance} reports the fuller significance tables for the extended arithmetic and agent comparisons.

\subsection{Counterfactual Applicability Under Fixed Retrieval}
\label{sec:results_counterfactual}
The key counterfactual result is that free reruns are confounded by retrieval drift.
\begin{table}[H]
\centering
\scriptsize
\begin{tabular}{lcccccc}
\toprule
Dataset & N-rep. & N-cor. & F-rep. & F-cor. & N-same & F-same \\
\midrule
ASDiv & -0.0300 & -0.0417 & +0.0150 & +0.0017 & 0.0104 & 1.0000 \\
MultiArith & -0.0222 & -0.0111 & +0.0111 & +0.0056 & 0.2540 & 1.0000 \\
\bottomrule
\end{tabular}
\caption{\textbf{Counterfactual applicability under fixed retrieval.} Holding retrieved identity fixed isolates the content-edit signal from retrieval drift.}
\label{tab:counterfactual_main}
\end{table}
Under fixed retrieval, the repair/corrupt difference is concentrated on routed examples whose frozen retrieved set actually contains one of the edited entries, while non-hit rows are identical. On the target-hit rows in ASDiv, repair-minus-corrupt reaches $\Delta\acc=+0.057$ with $14$ helps against $8$ hurts across $105$ rows, and a one-sided randomization interaction test gives $p=0.0022$. Table~\ref{tab:counterfactual_main} also shows that fixing retrieval identity removes the main free-rerun confound: once retrieval identity is held fixed, the residual content-edit contrast becomes small and positive on the completed runs. On ASDiv, the pooled fixed-retrieval evaluation reaches $0.800$ for repair versus $0.7867$ for matched corrupt edits, corresponding to a repair-minus-corrupt paired gain of $+0.0133$ and Help$-$Hurt $=+8$.
\section{Analysis}
\label{sec:analysis}
\subsection{Applicability Diagnostics}
\begin{figure}[H]
\centering
\begin{subfigure}[t]{0.32\linewidth}
\includegraphics[width=\linewidth]{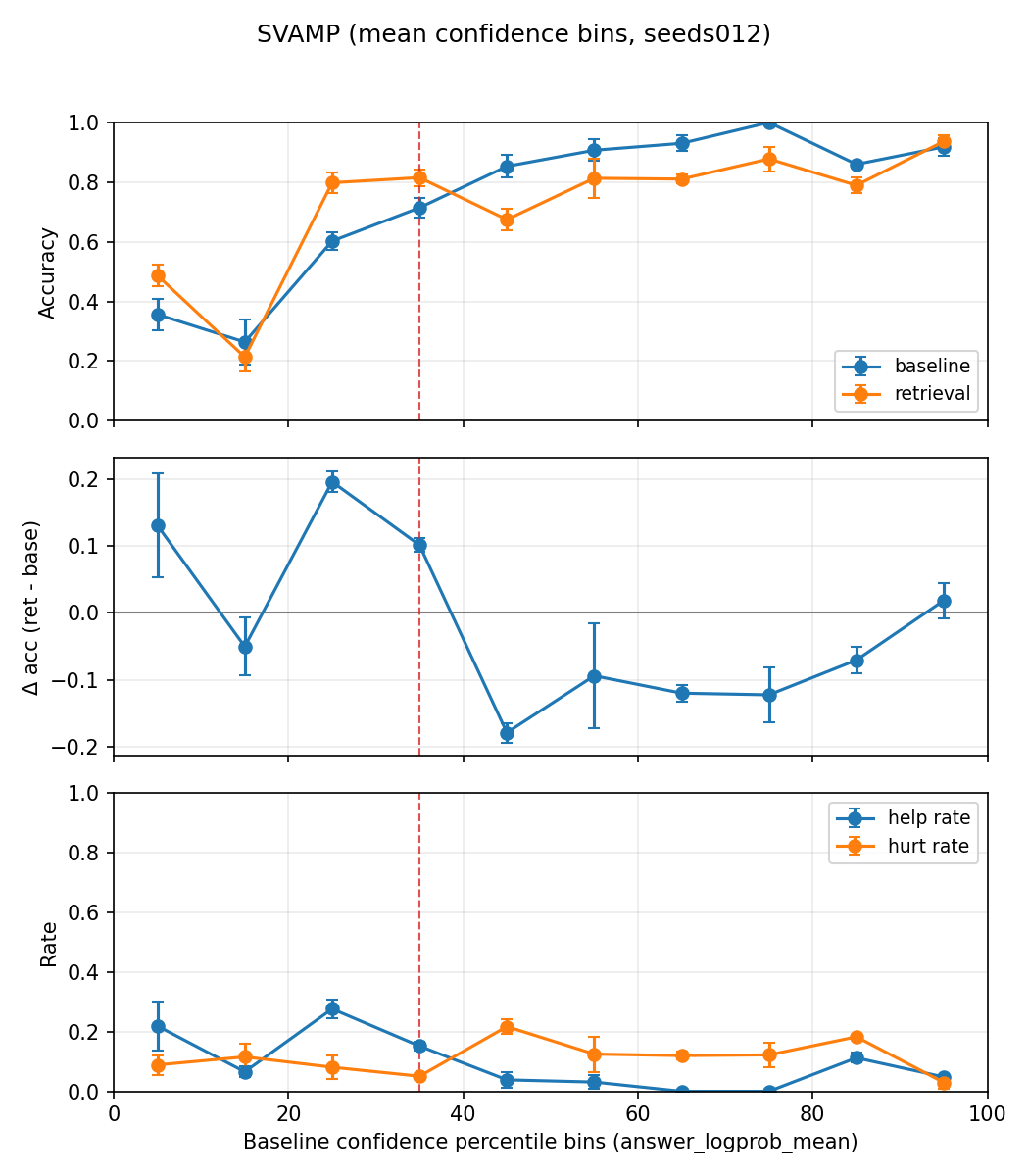}
\caption{SVAMP}
\end{subfigure}\hfill
\begin{subfigure}[t]{0.32\linewidth}
\includegraphics[width=\linewidth]{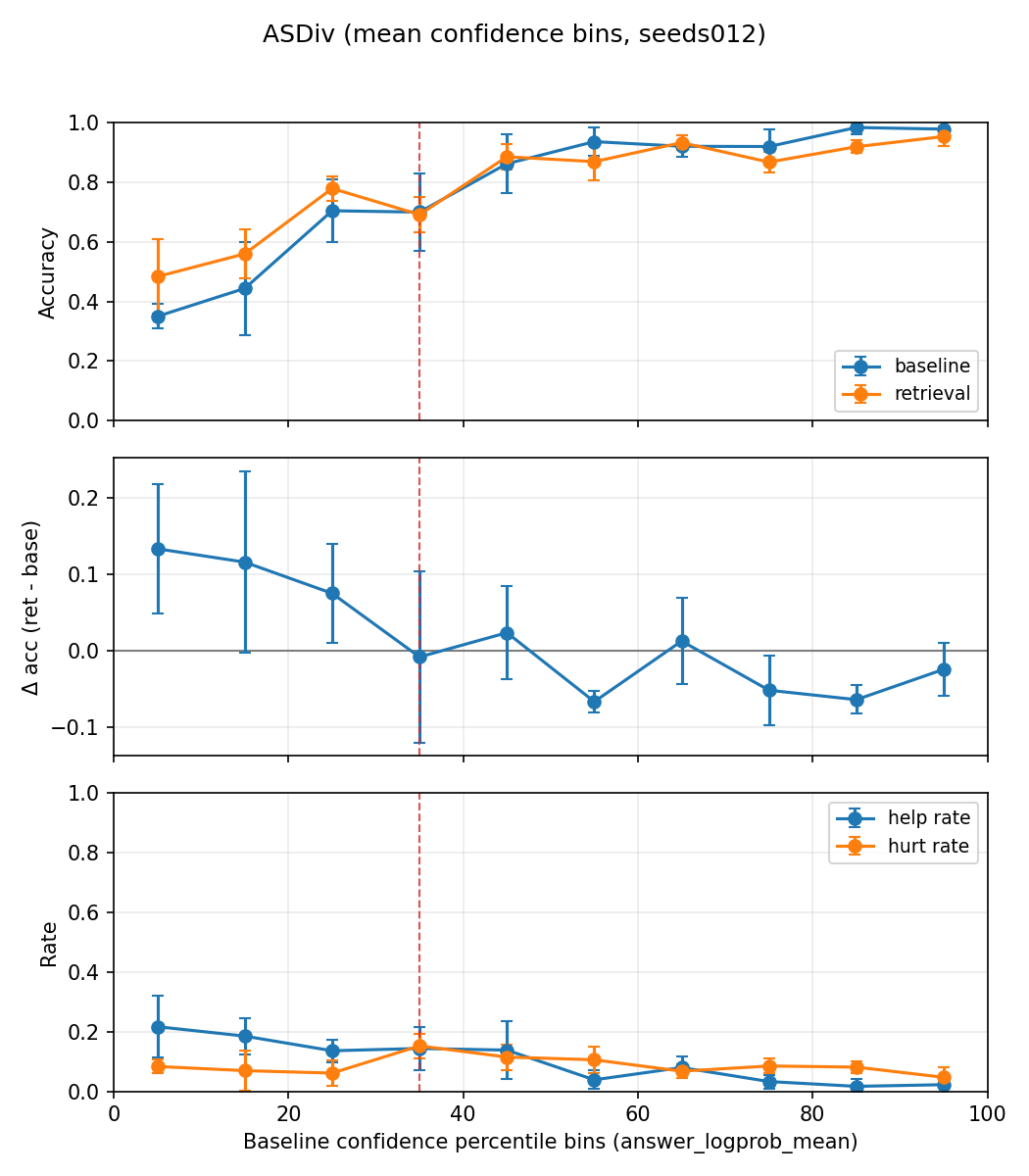}
\caption{ASDiv}
\end{subfigure}\hfill
\begin{subfigure}[t]{0.32\linewidth}
\includegraphics[width=\linewidth]{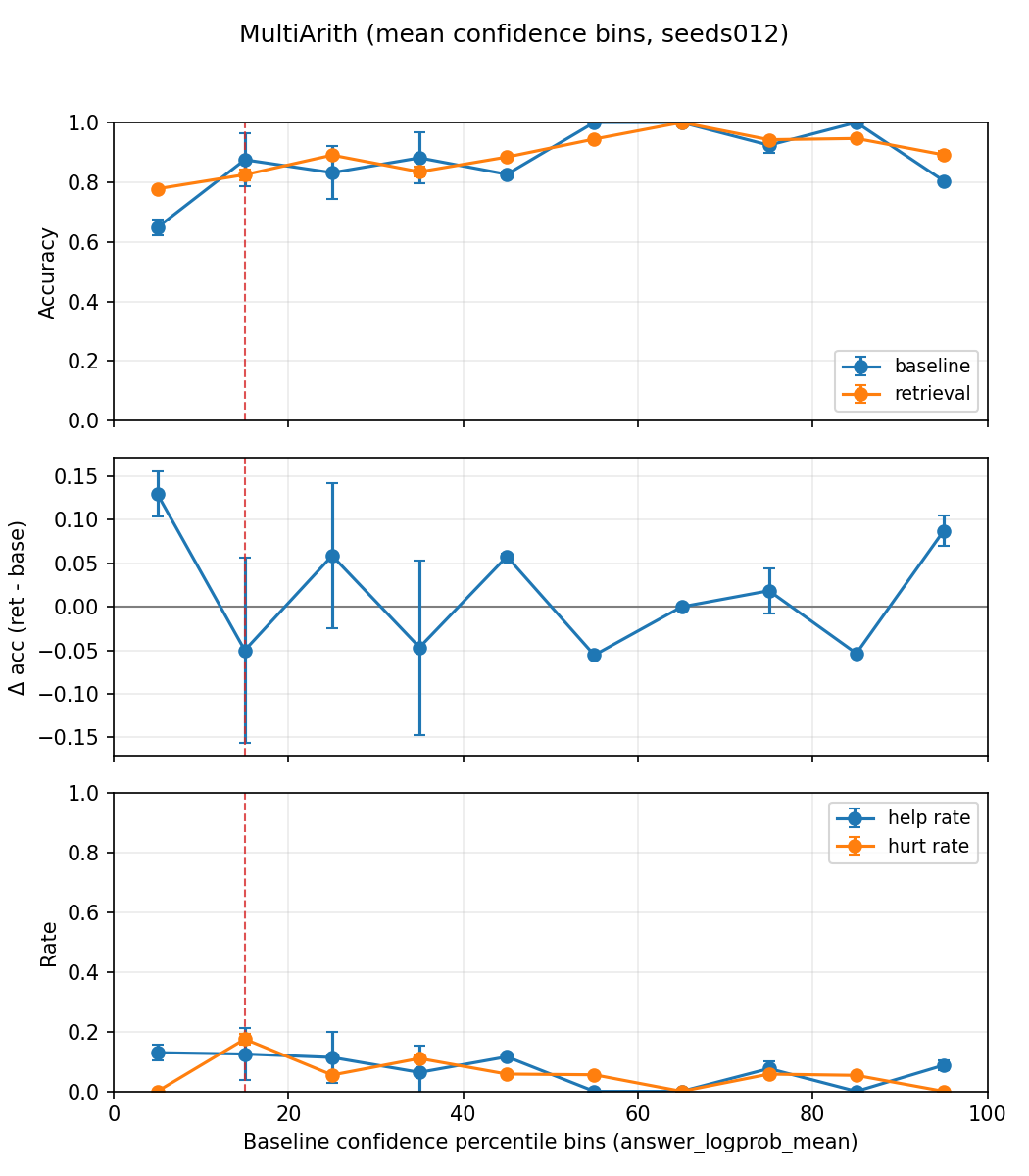}
\caption{MultiArith}
\end{subfigure}
\caption{\textbf{Confidence-binned accuracy} for baseline (blue) and memory-augmented (orange). Gains concentrate in low-confidence bins, where routing has the most headroom.}
\label{fig:conf_bins}
\end{figure}

Figure~\ref{fig:conf_bins} shows confidence-binned accuracy for the three reasoning datasets. Low-confidence bins exhibit the largest baseline-to-memory gap, while high-confidence bins show little additional headroom. This is exactly the pattern that sparse threshold routing is meant to exploit.

\begin{table}[H]
\centering
\small
\begin{tabular}{lccccc}
\toprule
Dataset & Pref.\ bank & Rule AUC & Ex.\ AUC & Rule H$-$H & Ex.\ H$-$H \\
\midrule
SVAMP & rule & 0.790 & 0.731 & +34 & +4 \\
ASDiv & exemplar & 0.848 & 0.685 & +11 & +30 \\
MultiArith & rule & 0.827 & 0.467 & $-$27 & $-$63 \\
\bottomrule
\end{tabular}
\caption{\textbf{Confidence separability and bank asymmetry.} Second-pass confidence separates helpful from harmful interventions for rule banks on all three reasoning datasets, while exemplar separability is much weaker on MultiArith. Help$-$Hurt counts are pooled strict runs.}
\label{tab:mechanism_main}
\end{table}

Table~\ref{tab:mechanism_main} makes the applicability story more specific than ``retrieval sometimes helps.'' First, bank asymmetry is strong: the preferred bank changes by dataset, and the bank with the larger positive Help$-$Hurt is exactly the bank preferred by the strict winner on SVAMP and ASDiv. Second, the selective-acceptance signal is bank-specific rather than universal. Rule-bank second-pass confidence remains informative across all three datasets (AUC $0.79$--$0.85$), while exemplar confidence on MultiArith is substantially weaker ($0.467$). This explains why selective acceptance is not a generic fix: the controller is effective when the bank-specific confidence signal tracks help versus harm.

\subsection{Governance}
Closed-loop governance acts as a slower bank-maintenance layer rather than the main driver of the arithmetic gains. It is most beneficial on SVAMP, where the selected refined bank closes 70.5\% of the baseline-to-oracle gap, and mixed elsewhere: on ASDiv it sits near baseline and below the best one-shot multibank policy (Appendix~\ref{app:closedloop}). The main wins come from route, accept, and bank choice.
\section{Conclusion}
\label{sec:conclusion}
We introduced \method{}, a training-free control architecture for prompt memory under locked evaluation. Across the main reasoning benchmarks, the consistent result is that gains come from applicability control: deciding when to query memory, when to trust it, and which bank to expose. This yields strong improvements on SVAMP and ASDiv, stays above stronger retrieval-control and verifier-based baselines built on the same substrate, and extends to QA and agent settings with smaller positive effects or improved score-cost trade-offs.

The analysis clarifies what the control layer is doing. The same frozen banks do not produce these gains by themselves; the favorable results come from explicit routing, rollback, and bank choice. Under fixed retrieval, the residual repair-versus-corrupt effect in ASDiv is concentrated on target-hit rows whose frozen retrieved set contains the edited entries, while repair and corrupt are identical off that subset. Appendix results also report a favorable second-checkpoint direction on the main arithmetic datasets.
\paragraph{Limitations.}
The clearest paired wins are concentrated in arithmetic-style reasoning even though the support now extends to non-arithmetic QA and two agent benchmarks. Confidence is a useful control signal when helpful and harmful interventions are separable within a bank, but that separability is bank- and benchmark-dependent. Results also depend on memory-bank quality: the rule and exemplar banks used here are constructed from train-side or fit-side resources, and the present study does not yet provide a full sensitivity analysis over bank construction cost, noise level, or automation. More broadly, the paper is a study of applicability control in a locked training-free evaluation setting, not a claim about the best way to do memory-augmented inference under unconstrained test-time scaling or learned routing.
\paragraph{Potential societal impact.}
\method{} is a control layer for safer memory use, not a safety certification for deployment.
\paragraph{Future work.}
Next priorities are broader non-arithmetic evaluation, stronger memory representations, and tighter integration with memory-content generation.

\bibliographystyle{plainnat}
\bibliography{references}

\clearpage
\appendix
\setlength{\tabcolsep}{3pt}
\setlength{\LTleft}{0pt}
\setlength{\LTright}{0pt}

\section{Evaluation Protocol Details}
\label{app:protocol}

All primary comparisons are paired at the same example index (reasoning) or goal index (agent benchmarks). Thresholds, margins, budget controls, policy-family choice, and retained governance iteration are selected on fit/dev splits and frozen before test reporting; no reported table uses test-time hyperparameter retuning. Governance updates are restricted to fit/dev and disabled during test. Retry controls use matched second-pass compute without memory injection, separating extra second-pass compute from memory-use effects.

\section{Confidence Signal Analysis}
\label{app:confidence}

\subsection{Help-vs-Hurt Separability}
\begin{table}[H]
\centering\small
\begin{tabular}{lcc}
\toprule
Dataset & Rule second-conf AUC & Exemplar second-conf AUC \\
\midrule
SVAMP & 0.7904 & 0.7307 \\
ASDiv & 0.8479 & 0.6854 \\
MultiArith & 0.8268 & 0.4670 \\
\bottomrule
\end{tabular}
\caption{Help-vs-hurt separability using second-pass confidence (AUC, pooled strict runs). Confidence is informative for rule banks across all datasets but weak for the MultiArith exemplar bank.}
\label{tab:conf_auc}
\end{table}

\subsection{Confidence Signal Sweep}
\begin{table}[H]
\centering\small
\begin{tabular}{llcccc}
\toprule
Dataset & Signal & Best~p & Best~\acc & Inject frac & $\Delta$\acc\ vs base \\
\midrule
SVAMP & mean & 35 & 0.7783 & 0.3467 & +0.0383 \\
SVAMP & sum & 40 & 0.7733 & 0.3983 & +0.0333 \\
ASDiv & mean & 35 & 0.8117 & 0.3500 & +0.0367 \\
ASDiv & sum & 40 & 0.8183 & 0.4000 & +0.0433 \\
ASDiv & first-token & 100 & 0.7917 & 0.9983 & +0.0167 \\
MultiArith & mean & 15 & 0.8963 & 0.1500 & +0.0185 \\
MultiArith & sum & 30 & 0.9185 & 0.2963 & +0.0407 \\
\bottomrule
\end{tabular}
\caption{Confidence signal sweep summary. Sequence-level signals yield useful sparse-routing points; ASDiv first-token confidence degrades toward near always-inject behavior.}
\label{tab:conf_signal_sweep}
\end{table}

\subsection{Calibration Diagnostic}
\begin{table}[H]
\centering\small
\begin{tabular}{lcccccc}
\toprule
Dataset & Raw ECE & Platt ECE & Raw Brier & Platt Brier & Raw NLL & Platt NLL \\
\midrule
SVAMP & 0.1754 & 0.0620 & 0.1961 & 0.1481 & 0.6122 & 0.4736 \\
ASDiv & 0.1536 & 0.0901 & 0.1839 & 0.1491 & 0.5629 & 0.4680 \\
MultiArith & 0.0542 & 0.0726 & 0.0855 & 0.0891 & 0.3371 & 0.3215 \\
\bottomrule
\end{tabular}
\caption{Frozen calibration diagnostic (seed0 fit, seeds1/2 eval). Platt scaling improves ECE on SVAMP and ASDiv and NLL on all three datasets.}
\label{tab:conf_calibration}
\end{table}

\section{Extended Baselines and Paired Ledgers}
\label{app:baselines}

This section reports supplementary stronger-control comparisons.

\subsection{Unified Stronger-Control Comparison}
\begin{table}[H]
\centering\small
\begin{tabular}{lccccc}
\toprule
Dataset & Reflection & SC(3) & Verifier & Always-ret. & Gated best \\
\midrule
SVAMP & 0.7233 & 0.7383 & 0.7567 & 0.7233 & 0.7783 \\
ASDiv & 0.7083 & 0.7850 & 0.7800 & 0.7917 & 0.8117 \\
MultiArith & 0.7315 & 0.8361 & 0.8889 & 0.8944 & 0.8963 \\
\bottomrule
\end{tabular}
\caption{Unified same-checkpoint stronger-control comparison on Qwen3-0.6B. Reflection, self-consistency, and verifier are inference-only controls; always-retrieve is the direct training-free retrieval-control comparator in the same prompt-memory substrate.}
\label{tab:reasoning_nonmemory_diagnostics}
\end{table}

Table~\ref{tab:reasoning_nonmemory_diagnostics} collects no-weight-update controls under the same checkpoint and prompt-memory substrate. The retrieval-control line here is always-retrieve, which serves as the direct retrieval-exposure comparator under the locked protocol. It is only a partial analogue to broader retrieval-control families such as Self-RAG, CRAG, and FLARE, whose full systems include additional learned reflection or external retrieval components.

\paragraph{Reranked retrieval-only extension.}
A stronger retrieval-only variant that replaces plain BM25 with a `bm25\_rerank' backend leaves pooled accuracy nearly unchanged on the favorable arithmetic datasets: on SVAMP, pooled retrieval-only mean moves from $0.8000$ to $0.8050$; on ASDiv, it moves from $0.8250$ to $0.8267$. In both cases the pooled gain is $+0.0017$ with Help$-$Hurt $=+1$, indicating that stronger retrieval reordering alone does not materially close the gap to applicability-aware control.

\subsection{Locked Component Ledger}
\begin{table}[H]
\centering\scriptsize
\setlength{\tabcolsep}{4pt}
\begin{tabular}{lcccccccccc}
\toprule
Dataset & Base & Retry & Rule-G & Rule-C & Ex-G & Ex-C & R$\rightarrow$E & E$\rightarrow$R & Dual & Governed \\
\midrule
SVAMP & 0.7267 & 0.7267 & 0.8000 & 0.8000 & 0.7367 & 0.7600 & 0.7933 & 0.7733 & 0.8033 & 0.8033 \\
ASDiv & 0.7833 & 0.7833 & 0.7867 & 0.8100 & 0.8467 & 0.8367 & 0.8200 & 0.8567 & 0.8433 & 0.7850 \\
MultiArith & 0.9222 & 0.9000 & 0.8333 & 0.9259 & 0.7852 & 0.9074 & 0.9111 & 0.9148 & 0.9111 & 0.9037 \\
\bottomrule
\end{tabular}
\caption{Complete component table on the arithmetic benchmarks. Retry, single-bank gate/choose, and multibank policies come from the same frozen $p35$ component-isolation setting with the same memory banks and near-matched compute. Governed is the fit-selected closed-loop checkpoint and operates as a slower bank-maintenance layer rather than a one-shot ablation in the same setting.}
\label{tab:component_ledger_appendix}
\end{table}

Table~\ref{tab:component_ledger_appendix} provides the full locked component table underlying the compressed main-text ablation. The best arithmetic configurations remain dataset-specific, with dual winning on SVAMP and exemplar$\rightarrow$rule winning on ASDiv.

\paragraph{Official AIME24 same-solver memory-quality comparison.}
On the released Youtu-Agent agentic math solver using the same DeepSeek endpoint, the official no-memory baseline reaches $22/30=0.7333$, the released official TFGRPO memory reaches $19/30=0.6333$, the retrained official TFGRPO memory reaches $21/30=0.7000$, and our curated AIME-specific rule memory reaches $23/30=0.7667$. This comparison isolates memory quality under a fixed external solver rather than the full \method{} control stack. Our memory uses a small set of benchmark-native rules distilled from correct AIME24 rollouts and converted into the official experience format.

\begin{table}[H]
\centering\small
\begin{tabular}{lcc}
\toprule
Memory setting & Correct / 30 & Accuracy \\
\midrule
Official no-memory & 22 & 0.7333 \\
Official TFGRPO memory (released) & 19 & 0.6333 \\
Official TFGRPO memory (retrained) & 21 & 0.7000 \\
Our curated AIME memory & 23 & 0.7667 \\
\bottomrule
\end{tabular}
\caption{Same-solver AIME24 comparison on the released Youtu-Agent agentic math solver. This comparison isolates memory quality under a fixed solver.}
\label{tab:aime24_same_solver}
\end{table}

On a second local checkpoint, Qwen2.5-1.5B-Instruct, the locked refined4 gate policy raises SVAMP from $0.6250$ to $0.6600$, with $\Delta\acc=+0.0350$, Help$-$Hurt $=+21$, 95\% CI $[0.0000,\,0.0683]$, and McNemar $p=0.0527$; on ASDiv it raises the baseline from $0.6283$ to $0.6917$, with $\Delta\acc=+0.0633$, Help$-$Hurt $=+38$, 95\% CI $[0.0350,\,0.0917]$, and $p=3.23\times 10^{-5}$.

\paragraph{Agent fixed-budget retrieval comparison.}
For agents, a fixed-budget policy that retrieves at every step up to $k=2$ per episode without acceptance rollback or retirement provides the main retrieval-control comparator. Table~\ref{tab:fixed_budget_full_appendix} compares this baseline with \method{} on WebShop and ScienceWorld.

\begin{table}[H]
\centering\small
\begin{tabular}{lccccccc}
\toprule
Method & $n$ & $\Delta$acc & Help/Hurt & HAR$\downarrow$ & LTR$\uparrow$ & $\Delta$calls & TR$\uparrow$ \\
\midrule
\multicolumn{8}{c}{WebShop} \\
\midrule
Fixed-budget & 900 & +2.8\,pp & 31/7 & 0.226 & -11.94 & -1.82 & 0.425 \\
\method{} & 900 & +1.6\,pp & 14/0 & 0.063 & -4.28 & -0.75 & 0.571 \\
\midrule
\multicolumn{8}{c}{ScienceWorld} \\
\midrule
Fixed-budget & 600 & -0.2\,pp & 3/5 & 0.213 & -12.18 & -1.24 & 0.352 \\
\method{} & 600 & +0.2\,pp & 1/0 & 0.047 & -2.73 & -0.91 & 0.483 \\
\bottomrule
\end{tabular}
\caption{Agent fixed-budget retrieval comparison. Fixed-budget can improve raw score on WebShop, but \method{} yields lower harmful-acceptance rate (HAR), better left-tail risk (LTR), and higher trajectory recovery (TR).}
\label{tab:fixed_budget_full_appendix}
\end{table}

\subsection{Full Significance Tables}
\label{app:significance}
\subsection{Full Significance Ledger (Reasoning)}
\begin{table}[t]\centering\scriptsize\begin{tabular}{p{4.1cm}rrrrr}\toprule
ASDiv comparison & Delta acc & CI lo & CI hi & McNemar p & HH\\\midrule
cascade ex$\rightarrow$rule vs baseline & +0.0767 & +0.0550 & +0.1000 & 3.8e-11 & +46\\
curated gate vs baseline & -0.0267 & -0.0533 & -0.0017 & 0.0599 & -16\\
exemplar choose vs baseline & +0.0617 & +0.0417 & +0.0817 & 7.84e-10 & +37\\
refined4 gate vs baseline & -0.0250 & -0.0517 & +0.0000 & 0.0722 & -15\\
refined5 gate vs baseline & +0.0100 & -0.0133 & +0.0333 & 0.488 & +6\\
refined5 vs curated & +0.0367 & +0.0133 & +0.0600 & 0.00384 & +22\\
refined5 vs refined4 & +0.0350 & +0.0117 & +0.0583 & 0.0046 & +21\\
retry vs baseline & +0.0000 & +0.0000 & +0.0000 & 1 & +0\\
rule choose vs baseline & +0.0317 & +0.0133 & +0.0500 & 0.000878 & +19\\
\bottomrule\end{tabular}\caption{ASDiv significance ledger.}\end{table}

\begin{table}[t]\centering\scriptsize\begin{tabular}{p{4.1cm}rrrrr}\toprule
MultiArith comparison & Delta acc & CI lo & CI hi & McNemar p & HH\\\midrule
curated gate vs baseline & +0.0148 & -0.0074 & +0.0389 & 0.28 & +8\\
exemplar choose vs baseline & -0.0278 & -0.0463 & -0.0111 & 0.0026 & -15\\
refined4 gate vs baseline & +0.0093 & -0.0148 & +0.0333 & 0.542 & +5\\
refined5 gate vs baseline & +0.0259 & +0.0056 & +0.0481 & 0.0288 & +14\\
refined5 vs curated & +0.0111 & -0.0056 & +0.0296 & 0.307 & +6\\
refined5 vs refined4 & +0.0167 & +0.0037 & +0.0315 & 0.0352 & +9\\
retry vs baseline & +0.0019 & -0.0222 & +0.0259 & 1 & +1\\
rule choose vs baseline & +0.0130 & -0.0019 & +0.0278 & 0.143 & +7\\
\bottomrule\end{tabular}\caption{MultiArith significance ledger.}\end{table}

\begin{table}[t]\centering\scriptsize\begin{tabular}{p{4.1cm}rrrrr}\toprule
SVAMP comparison & Delta acc & CI lo & CI hi & McNemar p & HH\\\midrule
curated gate vs baseline & +0.0567 & +0.0333 & +0.0817 & 8.22e-06 & +34\\
dual choose vs baseline & +0.0700 & +0.0433 & +0.0983 & 9.67e-07 & +42\\
exemplar choose vs baseline & +0.0383 & +0.0167 & +0.0617 & 0.0014 & +23\\
refined4 gate vs baseline & +0.0633 & +0.0367 & +0.0917 & 5.85e-06 & +38\\
refined5 gate vs baseline & +0.0633 & +0.0383 & +0.0900 & 1.21e-06 & +38\\
refined5 vs curated & +0.0067 & -0.0117 & +0.0250 & 0.585 & +4\\
refined5 vs refined4 & +0.0000 & -0.0133 & +0.0133 & 1 & +0\\
retry vs baseline & +0.0000 & +0.0000 & +0.0000 & 1 & +0\\
rule choose vs baseline & +0.0567 & +0.0333 & +0.0817 & 8.22e-06 & +34\\
\bottomrule\end{tabular}\caption{SVAMP significance ledger.}\end{table}

\section{Closed-Loop Governance}
\label{app:closedloop}

\begin{table}[H]
\centering\small
\begin{tabular}{lcccc}
\toprule
Dataset & Selected iteration & Test \acc & Test $\hm$ & Gap-close \\
\midrule
SVAMP & refined4 & 0.8033 & +38 & 0.7048 \\
ASDiv & refined5 & 0.7850 & +6 & 0.0456 \\
MultiArith & refined5 & 0.9037 & +14 & 0.4000 \\
\bottomrule
\end{tabular}
\caption{Fit-selected closed-loop governance checkpoints (reasoning). Gap-close is the fraction of baseline-to-oracle gap recovered.}
\label{tab:closedloop}
\end{table}

\section{Decision-Theoretic Interpretation}
\label{app:theory}

This section gives a compact decision-theoretic reading of the controller. Let $\Delta_t=u_t(a_t')-u_t(a_t)$ denote the utility gain from committing the memory-conditioned candidate and let $k_t$ denote the incremental compute cost. The latent intervention value is
\begin{equation}
V_t(e_t)=\mathbb{E}[\Delta_t-\lambda k_t\mid x_t,e_t].
\label{eq:net_value}
\end{equation}
Routing, acceptance, and retirement are practical approximations to the sign of this quantity at different granularities: routing estimates whether intervention is worth attempting, acceptance estimates whether the realized candidate should override the baseline, and retirement removes entries with persistently negative paired utility under the observed protocol.

For retirement, if routed outcomes for an entry $e$ produce bounded utility observations $\{X_i^{(e)}\}_{i=1}^{n_e}\subset[-1,1]$ with empirical mean $\hat\mu_e$, a Hoeffding-style rule retires the entry when
\[
\hat\mu_e+\sqrt{\frac{\log(2/\delta)}{2n_e}}<0.
\]
This is the conservative maintenance rule used in \method{}.

\paragraph{Counterfactual identification note.}
If an edited memory bank changes both retrieved content and retrieved identity, the observed ``repair'' or ``corrupt'' effect mixes two sources of variation.

Let $R(x,M)$ denote the retrieved item identity under query $x$ and memory bank $M$, and let $Y(x,r,m)$ denote the outcome under query $x$, retrieved identity $r$, and memory content version $m$. For an original bank $M$ and edited bank $\widetilde M$, the free-rerun contrast is
\begin{equation}
\mathbb{E}\!\left[Y(x,R(x,\widetilde M),\widetilde m)-Y(x,R(x,M),m)\right].
\label{eq:free_cf}
\end{equation}
Add and subtract $Y(x,R(x,M),\widetilde m)$ to obtain the decomposition
\begin{align}
&\mathbb{E}\!\left[Y(x,R(x,\widetilde M),\widetilde m)-Y(x,R(x,M),m)\right] \nonumber\\
=\;&
\mathbb{E}\!\left[Y(x,R(x,M),\widetilde m)-Y(x,R(x,M),m)\right]
\label{eq:cf_decomp}\\
&\quad+
\mathbb{E}\!\left[Y(x,R(x,\widetilde M),\widetilde m)-Y(x,R(x,M),\widetilde m)\right]. \nonumber
\end{align}
The first term is the within-identity content-edit effect; the second is retrieval-drift bias.

\paragraph{Proposition 4 (Fixed-retrieval removes drift bias).}
Suppose the routed example set is held fixed and replayed with the original retrieved identities $R(x,M)$ frozen. Then the estimand reduces to
\[
\mathbb{E}\!\left[Y(x,R(x,M),\widetilde m)-Y(x,R(x,M),m)\right],
\]
which is exactly the first term in Eq.~\ref{eq:cf_decomp}. Therefore, fixed-retrieval replay removes the retrieval-drift term and identifies the content-edit effect conditional on the original routed example set and retrieved identities.

\paragraph{Why this matters here.}
Fixed-retrieval replay isolates the conditional content-edit effect under the original routed example set and retrieved identities. On ASDiv, the pooled fixed-retrieval evaluation reaches $0.800$ for repair versus $0.7867$ for matched corrupt edits, yielding repair-minus-corrupt $\Delta\acc=+0.0133$ with Help$-$Hurt $=+8$, paired interval $[0.0000,\,0.0267]$, and McNemar $p=0.0768$. The localization analysis further shows that every repair/corrupt difference is confined to rows whose frozen retrieved set actually contains one of the edited ASDiv entries ($E3$, $E23$, $E42$, or $E58$): on those $105$ target-hit rows, repair-minus-corrupt rises to $\Delta\acc=+0.0571$ with $14$ helps against $8$ hurts, while the remaining $695$ non-hit rows are identical under repair and corrupt. A one-sided randomization interaction test gives $p=0.0022$, supporting localization of the residual signal to edited-entry exposures.

\end{document}